\documentclass[11pt]{article}
\usepackage[utf8]{inputenc}
\usepackage{amsmath,amssymb,amsfonts}
\usepackage{graphicx}
\usepackage{booktabs}
\usepackage{hyperref}
\usepackage{xcolor}
\usepackage{multirow}
\usepackage{array}
\usepackage{float}
\usepackage{enumitem}
\usepackage{caption}
\usepackage{wrapfig}
\usepackage{geometry}
\usepackage{rotating}

\title{\textbf{RFX-Fuse: Breiman and Cutler's Unified ML Engine + Native Explainable Similarity}\\[0.5em]}

\author{
    Chris Kuchar \\
    Independent Researcher \\
    \texttt{chrisjkuchar@gmail.com}
}

\date{\today}

\begin{document}

\maketitle

\vspace{-2em}

\begin{wrapfigure}{r}{0.45\textwidth}
\vspace{-1em}
\centering
\includegraphics[width=0.44\textwidth]{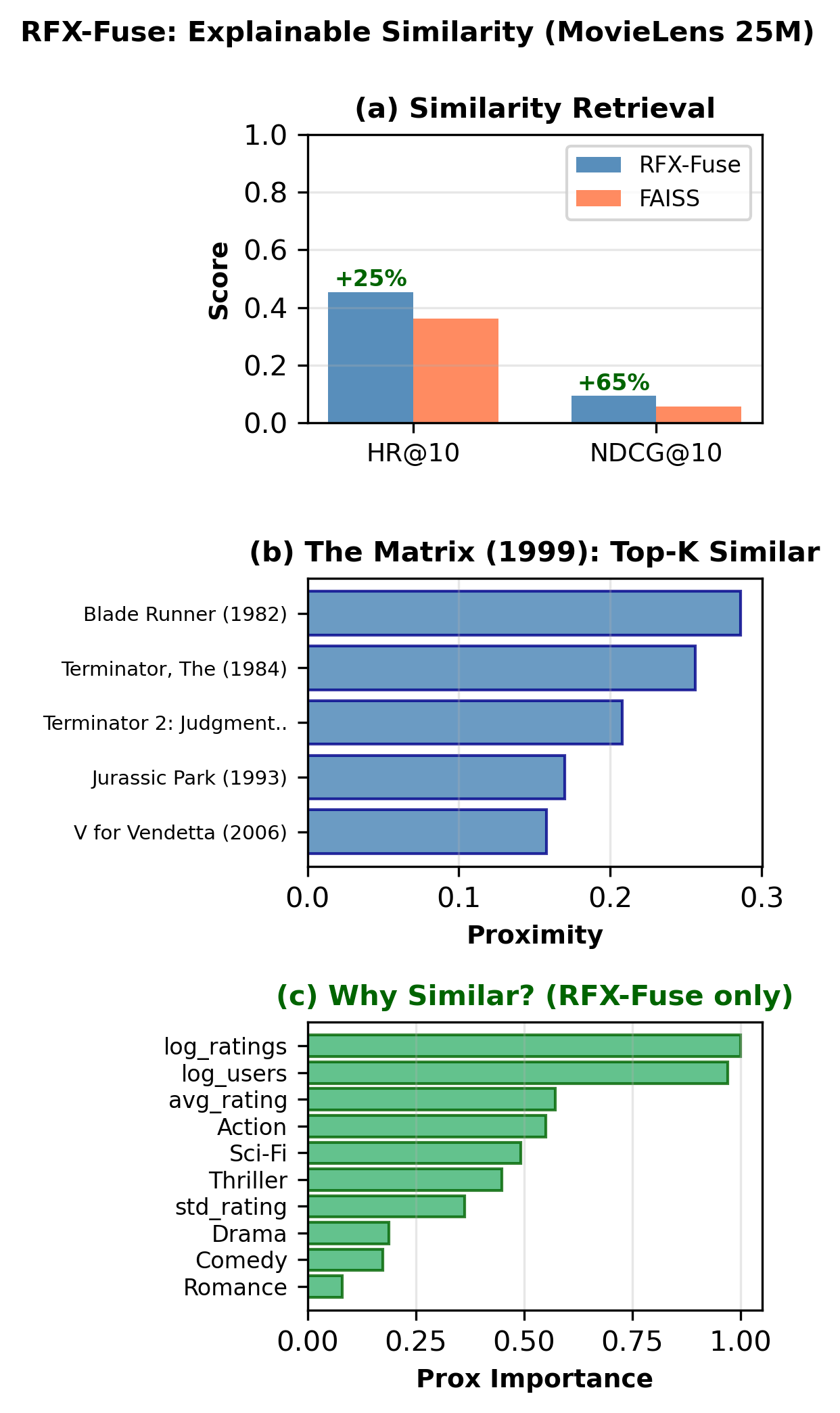}
\vspace{-1em}
\caption{\scriptsize Ground truth: user's liked items (rating $\geq$4.0) in held-out test set; 500 users. (a) RFX-Fuse vs FAISS. (b) The Matrix Top-K. (c) Why similar? (RFX-Fuse only*). \\ *\textit{RFX-Fuse only: native capability not available in compared tools (FAISS, XGBoost, sklearn RF, Isolation Forest).}}
\label{fig:hook}
\vspace{-1em}
\end{wrapfigure}

\noindent Breiman and Cutler's original Random Forest was designed as a unified ML engine---not merely an ensemble predictor. Their implementation included classification, regression, unsupervised learning, proximity-based similarity, outlier detection, missing value imputation, and visualization---capabilities that modern libraries like scikit-learn never implemented. RFX-Fuse (Random Forests X [X=compression]---Forest Unified Learning and Similarity Engine) delivers Breiman and Cutler's complete vision with native GPU/CPU support.

\vspace{0.4em}
\noindent Modern ML pipelines can require 5+ separate tools---XGBoost~\cite{xgboost2016} for prediction, FAISS~\cite{faiss2024} for similarity, SHAP~\cite{shap2017} for explanations, Isolation Forest~\cite{isolationforest2008} for outliers, custom code for importance. In most cases, a single RFX-Fuse model---a single set of trees grown once---provides all of these capabilities; in some cases two models can be used. We demonstrate RFX-Fuse as a comparable alternative to these tools across five industry use cases.

\vspace{0.4em}
\noindent Novel contributions include: (1) Proximity Importance---native explainable similarity: proximity measures \textit{that} samples are similar; proximity importance explains \textit{why}. (2) Dataset-specific imputation validation for general tabular data---ranking imputation methods by how ``real'' the imputed data looks, without ground truth labels.

\vspace{0.4em}
\noindent{\footnotesize
\begin{tabular}{@{}lcc@{}}
\textbf{Use Case} & \textbf{RFX-Fuse} & \textbf{Alternative to} \\
\hline
Recommender Systems & 1--2 models & 5 tools \\
Finance Explainability & 1 model & 3 tools \\
Time Series Regression & 1 model & 4 tools \\
Imputation Validation & 1 model & time series methods \\
Anomaly Detection & 1 model & 4 tools \\
\end{tabular}}

\vspace{0.5em}

\clearpage
\section{Introduction}

Dr. Breiman and Dr. Cutler designed Random Forest as a unified ML engine, not merely an ensemble prediction model. Their original implementation provided: predictions, similarity search, outlier detection, missing value imputation, and explainability---all from a single set of trees grown once. This unified design has profound implications for modern ML systems: where current architectures can require 5+ separate models (e.g., XGBoost for prediction, FAISS for similarity, SHAP for explanations, Isolation Forest for outliers, custom code for feature importance), Breiman and Cutler's vision enables one, and at most two, model object(s) as an alternative to entire multi-tool systems, vastly simplifying architecture development and deployment.

Breiman and Cutler's insight---that trees induce a \textit{data-adaptive metric}, a neighborhood structure, a similarity space via the proximity matrix---was ahead of its time. This is not KNN applied to forests; rather, proximity emerges naturally from how data groups and clusters as trees are grown---samples landing in the same terminal nodes share structure that the forest has learned. RFX-Fuse (Random Forests X [X=compression]---Forest Unified Learning and Similarity Engine) brings this vision to production scale via GPU acceleration, memory-efficient compression techniques (low-rank approximation, leaf assignment indexing, greedy outlier search), and a high-performance C++/CUDA backend. The result: Breiman and Cutler's unified ML engine on modern datasets with millions of samples, native sparse matrix support for recommender systems, and real-time similarity search with built-in explainability.

\subsection{Restored Features: Completing Breiman \& Cutler's Vision}

Breiman and Cutler's original Random Forest implementation included powerful capabilities that modern libraries like scikit-learn~\cite{sklearn2011} never implemented. RFX-Fuse restores these lost capabilities with modern GPU acceleration:

\begin{itemize}
    \item \textbf{Classification}: Breiman and Cutler's original Random Forest Classification---now with GPU/CPU dense/sparse support. \textit{Outputs:} predicted class labels, OOB error, overall/local feature importance, proximity matrix, overall/local proximity importance.

    \item \textbf{Regression}: Breiman and Cutler's original Random Forest Regression---now with GPU/CPU dense/sparse support. \textit{Outputs:} predicted values, OOB error, overall/local feature importance, proximity matrix, overall/local proximity importance.

    \item \textbf{Unsupervised Learning}: Breiman and Cutler's original mode that predicts real vs. synthetic data (creates synthetic labels internally from unlabeled input)---now with GPU/CPU dense/sparse support. \textit{Outputs:} predictions (real vs. synthetic), OOB error (real vs. synthetic classification accuracy), overall/local feature importance (which features distinguish real from noise), proximity matrix, overall/local proximity importance.

    \item \textbf{Prototypes}: Breiman and Cutler's original prototype identification---automatic discovery of representative samples that summarize clusters, useful for user segmentation and item categorization.

    \item \textbf{Outlier Detection}: Breiman and Cutler's original outlier measure---samples with low proximity to their own class are identified as outliers. For unsupervised mode, samples classified as ``real'' but with low proximity to other real samples are outliers---they have structure but don't belong to any cluster. Useful for fraud detection, data quality, and anomaly identification.

    \item \textbf{Visualization}: Breiman and Cutler's original visualization capabilities---3D MDS scaling, RFViz grid, and interactive feature discovery tools for exploring proximity structure and feature relationships.

    \item \textbf{Missing Value Imputation}: Breiman and Cutler's original proximity-weighted imputation method---filling missing values using weighted averages from similar samples (identified via proximity matrix). RFX-Fuse implements both the original Breiman-Cutler formulation and Young's 2017 refinement~\cite{young2017}.
\end{itemize}

\subsection{Novel Contributions}

Building on Breiman and Cutler's foundation, RFX-Fuse introduces two novel contributions:

\begin{itemize}
    \item \textbf{Proximity Importance}: A novel interpretability measure that explains \textit{why} items are similar. This answers the critical question: ``What features make these users/items alike?'' Available as both overall proximity importance (global similarity structure) and local proximity importance (why THIS sample is similar to its neighbors).

    \item \textbf{Unsupervised Imputation Quality Validation}: A novel application for general tabular data where RFX Unsupervised ranks imputation methods by how ``real'' the imputed data looks---without ground truth labels. Lower OOB error means the imputed values preserve the original dependency structure better.
\end{itemize}

\subsection{Extensions of Breiman \& Cutler's Original Capabilities}

RFX-Fuse extends Breiman and Cutler's proximity-based framework with efficient implementations for production use:

\begin{itemize}
    \item \textbf{Leaf Assignment Inverted Index}: A data structure that stores, for each terminal node across all trees, which training samples landed there. At inference time, a new sample traverses each tree once, looks up its leaf nodes in the inverted index, and retrieves similar items in O(1) per tree---enabling real-time similarity search with explanations.

    \item \textbf{Top-K Similar}: An API that returns the K most similar samples to a query, computed efficiently via the inverted index without materializing the full proximity matrix.

    \item \textbf{Top-K Similar with Explanations}: An API that returns similar samples \textit{along with feature-level explanations} of why they are similar---combining proximity lookup with proximity importance in a single call for production recommender systems.
\end{itemize}

\subsection{Performance Optimizations}

RFX-Fuse includes significant engineering optimizations for production-scale deployment:

\begin{itemize}
    \item \textbf{Native Sparse CSR Support}: First-class support for Compressed Sparse Row (CSR) matrices on both GPU and CPU, enabling training on Netflix-scale user-item matrices (100M+ users) without dense memory explosion.

    \item \textbf{GPU Acceleration}: Full GPU support for classification, regression, and unsupervised learning---both dense and sparse matrix inputs.

    \item \textbf{Split-Finding Strategies}: Two interchangeable approaches:
    \begin{itemize}
        \item \textbf{Histogram Binning}: O(bins) split finding instead of O(n log n) sorting, enabling 5--10$\times$ speedup on large datasets with $<$5\% accuracy trade-off
        \item \textbf{Pre-sorting}: One-time feature sorting with index reuse across tree growth for maximum accuracy on smaller datasets
    \end{itemize}

    \item \textbf{Batched GPU Training}: Amortized GPU memory allocation across tree batches, with intelligent memory tiering---shared memory for smaller data sizes (faster access), global memory for larger datasets (unlimited scale).
\end{itemize}

\subsection{RFX-Fuse Outputs: Four Types of Importance}

A single trained RFX-Fuse model provides four importance measures---two restored from Breiman \& Cutler, two novel:

\begin{table}[H]
\centering
\caption{Four Types of Importance from a Single RFX-Fuse Model}
\label{tab:unified}
\footnotesize
\begin{tabular}{l|l|l|c}
\toprule
\textbf{Type} & \textbf{Question Answered} & \textbf{Scope} & \textbf{Source} \\
\midrule
Overall Feature Importance & Which features drive predictions? & Global & Restored \\
Local Feature Importance & Why THIS prediction? & Per-sample & Restored \\
\midrule
Overall Proximity Importance & Which features define similarity? & Global & Novel \\
Local Proximity Importance & Why is THIS sample similar to neighbors? & Per-sample & Novel \\
\bottomrule
\end{tabular}
\end{table}

All four are available from Classification, Regression, and Unsupervised models. Demonstrated across recommender, classification, time series, imputation, and anomaly detection domains in Section~\ref{sec:applications}.

Table~\ref{tab:feature_comparison_v2} shows that among compared frameworks, RFX-Fuse is the only one providing prediction + similarity + explanations for both.

\begin{table}[H]
\centering
\caption{Model Capabilities Comparison: Use Case Features. Checkmarks ($\checkmark$) indicate full support, dashes (---) indicate absence, $\sim$ indicates partial support.}
\label{tab:feature_comparison_v2}
\footnotesize
\resizebox{\textwidth}{!}{%
\begin{tabular}{p{4.2cm}ccccc}
\toprule
\textbf{Feature} & \textbf{RFX-Fuse} & \textbf{RFX} & \textbf{XGBoost} & \textbf{sklearn RF} & \textbf{FAISS} \\
\midrule
\textbf{Core Learning Modes} & & & & & \\
Classification & $\checkmark$ & $\checkmark$ & $\checkmark$ & $\checkmark$ & --- \\
Regression & $\checkmark$ & --- & $\checkmark$ & $\checkmark$ & --- \\
Unsupervised & $\checkmark$ & --- & --- & --- & --- \\
\midrule
\textbf{Variable Importance} & & & & & \\
Overall importance & $\checkmark$ & $\checkmark$ & $\checkmark$ & $\checkmark$ & --- \\
Local importance (per-sample) & $\checkmark$ & $\checkmark$ & SHAP & --- & --- \\
\midrule
\textbf{Proximity/Similarity} & & & & & \\
Proximity/similarity scoring & $\checkmark$ & $\checkmark$ & --- & --- & $\checkmark$ \\
Overall proximity importance & $\checkmark$ & --- & --- & --- & --- \\
Local proximity importance & $\checkmark$ & --- & --- & --- & --- \\
Leaf assignment inverted index & $\checkmark$ & --- & --- & --- & $\checkmark$ \\
Top-K similar API & $\checkmark$ & --- & --- & --- & $\checkmark$ \\
Top-K similar with explanations & $\checkmark$ & --- & --- & --- & --- \\
Prototypes & $\checkmark$ & --- & --- & --- & --- \\
\midrule
\textbf{Outlier Detection} & & & & & \\
Breiman-Cutler outlier scores & $\checkmark$ & $\checkmark$ & --- & --- & --- \\
Full mode (exact) & $\checkmark$ & $\checkmark$ & --- & --- & --- \\
Greedy mode (fast, $\sim$98\% correlation) & $\checkmark$ & --- & --- & --- & --- \\
Outlier detection with explanations & $\checkmark$ & --- & $\sim$ (3 models + SHAP) & --- & --- \\
\midrule
\textbf{Data Imputation} & & & & & \\
Proximity-weighted imputation (Breiman-Cutler) & $\checkmark$ & --- & --- & --- & --- \\
OOB-based imputation (Young-Cutler 2017) & $\checkmark$ & --- & --- & --- & --- \\
Unsupervised imputation quality validation & $\checkmark$ & --- & --- & --- & --- \\
Rank imputation methods (no ground truth) & $\checkmark$ & --- & --- & --- & --- \\
\bottomrule
\end{tabular}%
}
\end{table}

\section{Method}

For detailed algorithmic formulas, see the RFX technical paper~\cite{rfx_v1}. This section provides an architecture overview and details RFX-Fuse's key capabilities: proximity computation, proximity importance, outlier detection, sparse support, missing value imputation, and cold start handling.

\subsection{Architecture Overview}

RFX-Fuse provides three complementary learning modes, each with built-in similarity and explainability:

\begin{itemize}
    \item \textbf{Classification}: Predicts class labels with OOB error estimation. Outputs: predictions, overall/local feature importance, proximity matrix, overall/local proximity importance.

    \item \textbf{Regression}: Predicts continuous values with OOB error estimation. Outputs: predictions, overall/local feature importance, proximity matrix, overall/local proximity importance.

    \item \textbf{Unsupervised}: Predicts real vs. synthetic data (creates synthetic labels internally from unlabeled input), with OOB error estimation. Outputs: predictions (real vs. synthetic), overall/local feature importance (which features distinguish real from noise), proximity matrix, overall/local proximity importance. Uses Breiman and Cutler's original approach~\cite{breiman_cutler_rf}:

    \begin{quote}
    \textit{``The approach in random forests is to consider the original data as class 1 and to create a synthetic second class of the same size that will be labeled as class 2. The synthetic second class is created by sampling at random from the univariate distributions of the original data... Thus, class 2 has the distribution of independent random variables, each one having the same univariate distribution as the corresponding variable in the original data. Class 2 thus destroys the dependency structure in the original data. But now, there are two classes and this artificial two-class problem can be run through random forests.''}

    \textit{``If the OOB misclassification rate in the two-class problem is, say, 40\% or more, it implies that the x-variables look too much like independent variables to random forests. The dependencies do not have a large role. If the misclassification rate is lower, then the dependencies are playing an important role.''}

    \textit{``Formulating it as a two class problem has a number of payoffs. Missing values can be replaced effectively. Outliers can be found. Variable importance can be measured. Scaling can be performed. But the most important payoff is the possibility of clustering.''}
    \end{quote}

    \textbf{Implementation}: Fisher-Yates shuffle~\cite{fisher_yates1938} for synthetic generation, following Breiman and Cutler's original Fortran.
\end{itemize}

All three modes support GPU and CPU execution with native dense and sparse (CSR) matrix inputs. The key insight is that all three modes produce proximity, enabling similarity search from any supervised or unsupervised model.

\subsection{Prototypes}

Breiman and Cutler's original prototype identification provides representative samples that summarize clusters~\cite{breiman_cutler_rf}:

\begin{quote}
\textit{``Prototypes are a way of getting a picture of how the variables relate to the classification. For the jth class, we find the case that has the largest number of class j cases among its k nearest neighbors, determined using the proximities. Among these k cases we find the median, 25th percentile, and 75th percentile for each variable. The medians are the prototype for class j and the quartiles give an estimate of its stability.''}

\textit{``For the second prototype, we repeat the procedure but only consider cases that are not among the original k, and so on.''}
\end{quote}

Formally, for class $c$, the first prototype is:
\begin{equation}
    \text{prototype}_c^{(1)} = \arg\max_{i \in \text{class } c} \left| \{j : j \in \text{kNN}(i) \cap \text{class } c\} \right|
\end{equation}

Prototypes are useful for user segmentation (``representative users of each segment'') and item categorization (``canonical examples of each genre'').

\subsection{Proximity Importance}

Proximity tells us \textit{who} is similar; proximity importance tells us \textit{why}. This paper introduces a novel measure quantifying which features drive similarity structure. Existing similarity methods---FAISS~\cite{faiss2024}, cosine similarity, embedding-based approaches---return scores but provide no feature-level explanation of \textit{why} samples are similar. Interpretability tools like SHAP~\cite{shap2017} and LIME explain predictions, not pairwise similarity. SPINEX~\cite{spinex2023} explains predictions via neighbor analysis; Proximity Importance explains the similarity itself---which features cause samples to co-locate in terminal nodes.

\textbf{Definition.} For sample $i$ and feature $k$, the proximity importance $\Pi_{ik}$ is the fraction of correctly-predicted OOB trees where permuting feature $k$ changes sample $i$'s terminal node:
\[
\Pi_{ik} = \frac{1}{|\mathcal{T}_i^{\text{correct}}|} \sum_{t \in \mathcal{T}_i^{\text{correct}}} \mathbf{1}\left[\ell_t(x_i) \neq \ell_t(\tilde{x}_i^{(k)})\right]
\]
where $\mathcal{T}_i^{\text{correct}}$ is the set of trees where $i$ is OOB and correctly predicted, $\ell_t(\cdot)$ returns the terminal node in tree $t$, and $\tilde{x}_i^{(k)}$ is $x_i$ with feature $k$ replaced by a random donor's value.

\textbf{Computation.} Following the same approach as Breiman and Cutler's local variable importance~\cite{breiman_cutler_rf}: for each tree $t$ and each OOB sample $i$ where the prediction is correct, traverse tree with each feature $k$ permuted; if terminal node differs from original, accumulate $1/|\text{OOB}_t^{\text{correct}}|$. Both measures use identical normalization: accumulate $1/|\text{OOB}_t^{\text{correct}}|$ per tree, resulting in values in $[0,1]$. The key difference is what we measure: local variable importance measures prediction accuracy change; proximity importance measures terminal node change. Same sampling strategy and normalization, different signal.

\textbf{Interpretation.} $\Pi_{ik} \in [0,1]$ represents feature $k$'s contribution to sample $i$'s similarity structure. Values are rankings: $\Pi_{ik} = 0.65$ indicates feature $k$ is more important than a feature with $\Pi_{ik} = 0.40$ for sample $i$'s placement in the forest.

\textbf{Interpretation by mode:}
\begin{itemize}
    \item \textbf{Unsupervised}: No target needed. The real vs. synthetic approach groups samples that share dependency structure (``real'' data clusters together). Proximity importance explains \textit{why samples look real together}---which features they share that distinguish them from random noise and other groups of real data. Example: ``Similar because: Animation=0.44, Adventure=0.48''

    \item \textbf{Classification/Regression}: Similarity is \textit{with respect to the prediction task}. Samples group in terminal nodes because they share similar feature content that drives the prediction. Proximity importance explains which features cause samples to receive similar predictions. Example: ``Similar risk profile because: num\_medications=0.089, prior\_admissions=0.076''
\end{itemize}

\subsection{Outlier Detection}

Breiman and Cutler~\cite{breiman_cutler_rf} defined outliers as samples with low proximity to their own class. RFX-Fuse implements their original formula and an efficient greedy approximation for large-scale deployment.

\subsubsection{Breiman-Cutler Formula}

For sample $n$ in class $j$, define the average squared proximity to other class members:
\[
\bar{p}_n = \frac{1}{N_j - 1} \sum_{m \in \text{class } j, m \neq n} \text{prox}(n, m)^2
\]

The \textbf{raw outlier measure} is:
\[
\text{raw}_n = \frac{N_j}{\sum_{m \in \text{class } j, m \neq n} \text{prox}(n, m)^2}
\]

The \textbf{normalized outlier score} uses median absolute deviation (MAD) within each class:
\[
\text{outlier}_n = \frac{\text{raw}_n - \text{median}_j(\text{raw})}{\text{MAD}_j(\text{raw})}
\]

\textbf{Interpretation:} Scores $> 10$ typically indicate outliers. For regression, all samples are treated as one class.

\textbf{Contrast with Isolation Forest:} Isolation Forest~\cite{isolationforest2008} identifies outliers via \textit{path length}---samples that require fewer splits to isolate are outliers. This detects ``easy to separate'' points with extreme marginal values. Breiman-Cutler outlier detection identifies samples with \textit{low proximity} to others in the forest-induced similarity space---samples that don't belong to any learned cluster. These approaches are complementary: IF finds fast-split outliers (extreme individual features); RFX-Fuse finds manifold outliers (unusual feature \textit{combinations} that don't cluster with any group).

\subsection{Native Sparse Support}

User-item matrices are extremely sparse (typically $<1\%$ non-zero). RFX-Fuse implements native CSR (Compressed Sparse Row) support that:
\begin{itemize}
    \item Stores only non-zero values: $O(\text{nnz})$ memory
    \item Accesses features via sparse lookup during tree growing
    \item Avoids dense conversion that would require $O(n \times m)$ memory
\end{itemize}

\subsection{Missing Value Imputation}

RFX-Fuse implements two proximity-based imputation methods from the Random Forest literature.

\subsubsection{Breiman-Cutler Original Imputation}

Breiman and Cutler's original approach~\cite{breiman_cutler_rf} uses iterative proximity-weighted imputation:

\begin{enumerate}
    \item \textbf{Initial imputation}: For continuous variables, impute with median; for categorical, impute with mode.
    \item \textbf{Train forest}: Grow Random Forest on the imputed data.
    \item \textbf{Compute proximities}: Build $N \times N$ proximity matrix from terminal node co-occurrence.
    \item \textbf{Re-impute}: For missing value $x_{ik}$ (sample $i$, feature $k$):
    \begin{itemize}
        \item \textbf{Continuous}: Proximity-weighted average:
        \[
        \hat{x}_{ik} = \frac{\sum_{j: x_{jk} \text{ observed}} \text{prox}(i,j) \cdot x_{jk}}{\sum_{j: x_{jk} \text{ observed}} \text{prox}(i,j)}
        \]
        \item \textbf{Categorical}: Proximity-weighted mode:
        \[
        \hat{x}_{ik} = \arg\max_c \sum_{j: x_{jk} = c} \text{prox}(i,j)
        \]
    \end{itemize}
    \item \textbf{Iterate}: Repeat steps 2--4 until convergence (typically 4--6 iterations).
\end{enumerate}

\subsubsection{Young-Cutler 2017 Refinement}

Young~\cite{young2017} refined the original method to better handle heavy missingness by incorporating OOB predictions:

\begin{enumerate}
    \item \textbf{Initial imputation}: Same as Breiman-Cutler (median/mode).
    \item \textbf{Train forest}: Grow Random Forest on imputed data.
    \item \textbf{OOB-based re-imputation}: For missing $x_{ik}$ where sample $i$ is OOB in tree $t$:
    \[
    \hat{x}_{ik}^{(t)} = \frac{\sum_{j \in \text{leaf}_t(i), j \neq i, x_{jk} \text{ observed}} x_{jk}}{|\{j \in \text{leaf}_t(i), j \neq i, x_{jk} \text{ observed}\}|}
    \]
    \item \textbf{Aggregate across OOB trees}: Final imputation is the average of OOB-based estimates:
    \[
    \hat{x}_{ik} = \frac{1}{|\mathcal{T}_i^{\text{OOB}}|} \sum_{t \in \mathcal{T}_i^{\text{OOB}}} \hat{x}_{ik}^{(t)}
    \]
    where $\mathcal{T}_i^{\text{OOB}}$ is the set of trees where sample $i$ was out-of-bag.
    \item \textbf{Iterate}: Repeat until convergence.
\end{enumerate}

\textbf{Key difference}: Young's method uses only OOB terminal node members for imputation, avoiding information leakage from in-bag samples and providing more robust estimates when missingness is high ($>$30\%).

\subsection{Leaf Assignment Inverted Index}

A key bottleneck for proximity-based similarity search is the $O(n^2)$ proximity matrix. RFX-Fuse introduces a leaf assignment inverted index that eliminates this requirement:

\begin{verbatim}
Data Structure: inverted_index[tree_id][leaf_id] -> list of sample_ids

Query Algorithm (O(T x avg_leaf_size)):
  1. new_sample traverses each tree -> get leaf_ids
  2. For each leaf_id: lookup inverted_index -> get co-located samples
  3. Count sample occurrences across trees
  4. Return top-K by count (= proximity score)
\end{verbatim}

This enables real-time similarity queries without materializing the full proximity matrix.

\subsection{Cold Start Handling}

Unlike embedding-based systems that require interaction history, RFX-Fuse handles cold starts naturally through feature-based generalization:

\begin{itemize}
    \item \textbf{New Item/Patient/Borrower}: Add content features (metadata, demographics, attributes) $\rightarrow$ RF model immediately places it in proximity space $\rightarrow$ \texttt{get\_top\_k\_similar()} returns similar samples with explanations
    \item \textbf{New User}: Use stated preferences or demographics $\rightarrow$ RF Regressor/Classifier \texttt{predict()} returns predictions with local importance explaining each decision
\end{itemize}

No heuristics, no ``show popular items'' fallback. The model generalizes from features, not memorized interactions---and explains every decision.

%==============================================================================
\section{Industry Use Cases}
\label{sec:applications}
%==============================================================================

RFX-Fuse is evaluated across five use cases:

\vspace{0.5em}
\noindent
{\small
\begin{tabular}{@{}ll@{}}
\textbf{Use Case 1:} & Recommender Systems --- similarity + ranking with explanations \\
\textbf{Use Case 2:} & Finance Explainability --- complete 4-type explainability for regulatory compliance \\
\textbf{Use Case 3:} & Time Series Regression --- prediction, similarity, and imputation in one model \\
\textbf{Use Case 4:} & Imputation Validation --- ranking imputation methods without ground truth \\
\textbf{Use Case 5:} & Anomaly Detection --- identifying outliers via synthetic discrimination \\
\end{tabular}
}

\subsection{Benchmark Environment and Runtimes}

All experiments were conducted on consumer-grade hardware to demonstrate RFX-Fuse's accessibility:

\begin{table}[H]
\centering
\footnotesize
\begin{tabular}{@{}ll@{}}
\toprule
\textbf{Component} & \textbf{Specification} \\
\midrule
GPU & NVIDIA GeForce RTX 3060 (12GB VRAM) \\
CPU & AMD Ryzen 7 5800X (8 cores, 16 threads) \\
RAM & 32GB DDR4 \\
OS & Ubuntu 22.04 (WSL2) \\
RFX-Fuse & v1.0.0 (CUDA 12.8) \\
\bottomrule
\end{tabular}
\caption*{\textbf{Hardware Configuration}}
\end{table}

\begin{table}[H]
\centering
\footnotesize
\begin{tabular}{@{}l r r r r@{}}
\toprule
\textbf{Use Case} & \textbf{Train Size} & \textbf{Features} & \textbf{Trees} & \textbf{Training Time} \\
\midrule
1. Recommender (Unsup) & 59,047 ($\times$2) & 23 & 1,000 & 1,254s \\
1. Recommender (Sup) & 47,237 & 21 & 1,000 & 120.1s \\
2. Finance Classification & 46,396 & 15 & 500 & 69.0s \\
3. Bike Regression & 5,725 & 4 & 1,000 & 24.0s \\
4. Imputation Validation & 3,000 & 12 & 100 & 3.6s \\
5. Anomaly Detection & 15,000 & 8 & 100 & 111.5s \\
\bottomrule
\end{tabular}
\caption*{\textbf{Training Runtimes by Use Case} (GPU-accelerated)}
\end{table}

\noindent\textit{Note: Training times include proximity importance, local importance, and leaf assignment computation. All models trained with full explainability features enabled. Unsupervised models double the effective training size (original + synthetic permuted data).}

\vspace{0.5em}
\noindent\textbf{Tree Count Selection:} The table above shows RFX-Fuse using 100--1,000 trees depending on dataset size. In XGBoost comparisons, we use early stopping (typically 50--200 trees). This difference reflects a fundamental distinction between Random Forests and boosting:
\begin{itemize}
    \item \textbf{Random Forests}: Each tree is grown fully independent to minimum node size, enabling proximity-based explainability throughout (samples landing in the same terminal nodes share learned structure). More trees = more averaging = lower variance. RF \textit{cannot overfit} by adding trees---performance monotonically improves or plateaus.
    \item \textbf{XGBoost (Boosting)}: Trees are shallow and sequential (each corrects prior errors), optimized for fast prediction. More trees = sequential error correction = \textit{can overfit}. Requires early stopping or regularization to prevent overfitting.
\end{itemize}
\noindent XGBoost is faster for prediction; RFX-Fuse is slower on initial training due to growing full independent trees to minimum node size. This deeper tree structure is precisely what enables RFX-Fuse's proximity matrix and explainability features. Both methods are configured at their respective optimal settings: RFX-Fuse with sufficient trees for convergence, XGBoost with early stopping on a validation set.

\vspace{0.5em}
\noindent\textbf{Training Data Efficiency:} RFX-Fuse trains on 100\% of available training data because Out-of-Bag (OOB) error provides built-in validation---each tree is validated on the $\sim$37\% of samples it did not see during bootstrap sampling. XGBoost, to prevent overfitting, uses 20\% of training data for early stopping validation. This is a structural advantage of Random Forests: RFX-Fuse uses all data for learning while still obtaining unbiased error estimates. For time-series data (Use Case 3), XGBoost's validation set is the final temporal portion of training data to preserve chronological integrity.

%==============================================================================
\subsection{Use Case 1: Recommender Systems}
%==============================================================================

\begin{table}[H]
\centering
\footnotesize
\begin{tabular}{@{}l p{12cm}@{}}
\toprule
\textbf{Aspect} & \textbf{Details} \\
\midrule
Problem & Content-based similarity search and ranking with explanations. \\
Dataset & MovieLens 25M~\cite{movielens2015} (59,047 movies, 25M ratings, 23 item features: 19 genres + 4 rating statistics). Models trained on item features. \\
\midrule
RFX-Fuse Pipeline & RFX-Fuse Unsupervised (retrieval) $\rightarrow$ RFX-Fuse Supervised (re-ranking). \\
Industry Pipeline & FAISS (retrieval) $\rightarrow$ XGBoost (ranking) + SHAP (explanations) + Isolation Forest (outliers) + custom code. \\
\midrule
Evaluation & 500 users (claim 7 style~\cite{movielens2015}); ground truth = liked items (rating $\geq$4.0) in held-out test set; NDCG@10, HR@10. \\
\bottomrule
\end{tabular}
\caption*{\textbf{Use Case 1 Summary}}
\end{table}

\begin{table}[H]
\centering
\footnotesize
\begin{tabular}{@{}l l p{9cm}@{}}
\toprule
\textbf{Stage} & \textbf{Comparison} & \textbf{Evaluation Method} \\
\midrule
\multicolumn{3}{l}{\textit{Stage 1: Retrieval (Unsupervised)}} \\
\midrule
Similarity & RFX-Fuse vs FAISS & NDCG@10, HR@10 using ground truth: user's liked items (rating $\geq$4.0) in held-out test set. \\
Explainability & RFX-Fuse vs FAISS & RFX-Fuse provides overall + local proximity importance (RFX-Fuse only). \\
\midrule
\multicolumn{3}{l}{\textit{Stage 2: Ranking (Supervised) --- can serve as complete pipeline if content-based retrieval not required}} \\
\midrule
Prediction & RFX-Fuse vs XGBoost & RMSE on held-out test set (rating prediction). \\
Overall Var Imp & RFX-Fuse vs XGBoost & Both provide global feature rankings; RFX-Fuse uses split-based importance. \\
Local Var Imp & RFX-Fuse vs SHAP & Per-sample feature contributions explaining individual predictions. \\
Similarity & RFX-Fuse vs FAISS & Re-ranking Top-K using prediction-space proximity vs cosine similarity. \\
Outlier Detection & RFX-Fuse vs Isolation Forest & Outlier overlap, score correlation, and interpretability (RFX-Fuse explains \textit{why} via proximity importance). \\
\bottomrule
\end{tabular}
\caption*{\textbf{Evaluation Methodology}}
\end{table}

%--- FIGURE 1: RFX Unsupervised vs FAISS ---
\begin{figure}[H]
\centering
\includegraphics[width=\textwidth]{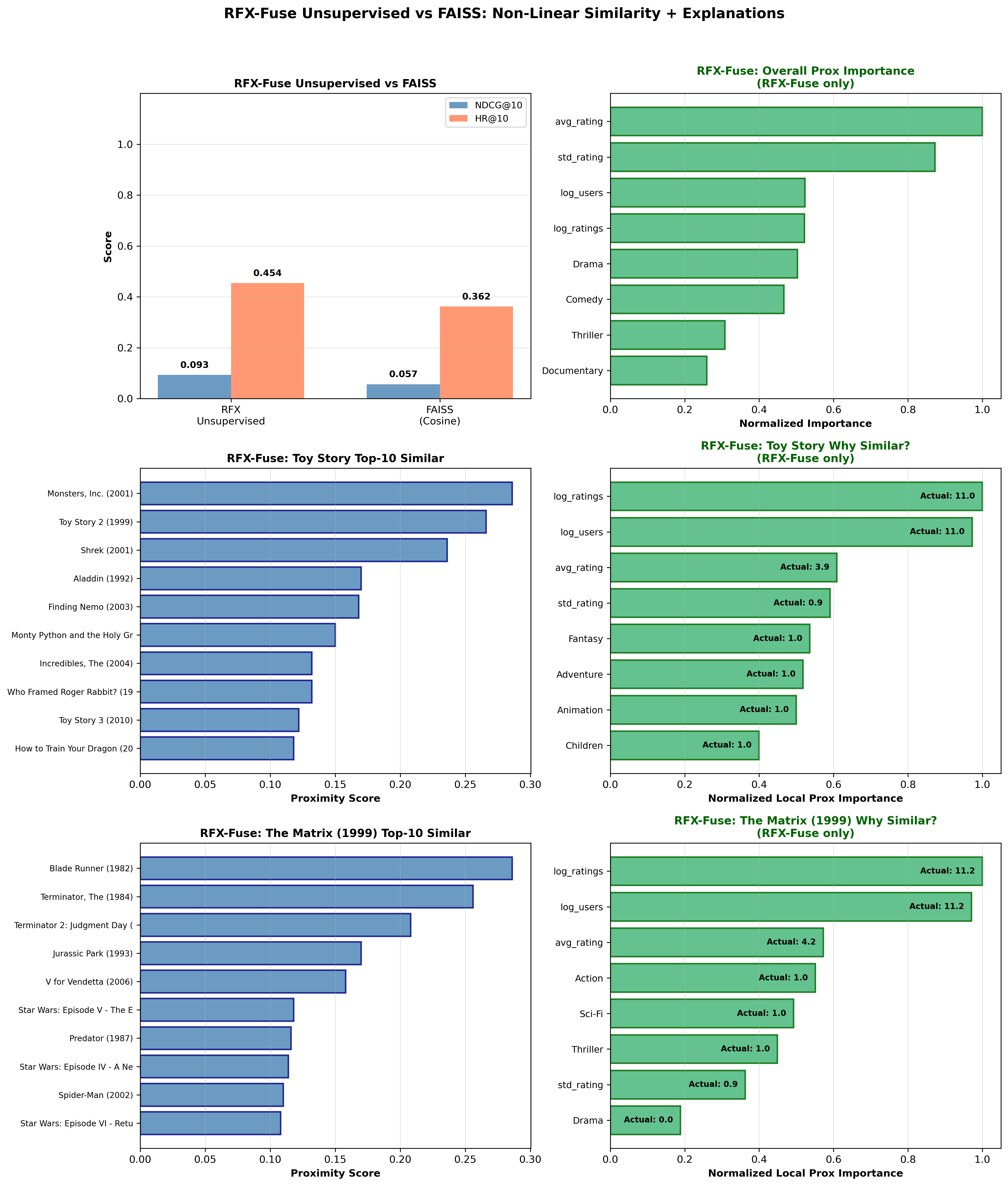}
\caption{RFX-Fuse Unsupervised vs FAISS for similarity retrieval.}
\label{fig:uc1-unsup}
\end{figure}

\begin{table}[H]
\centering
\footnotesize
\begin{tabular}{@{}cl p{10.5cm}@{}}
\toprule
\textbf{Panel} & \textbf{Title} & \textbf{Description} \\
\midrule
(a) & Metrics & RFX-Fuse achieves +64.6\% NDCG@10 and +25.4\% HR@10 vs FAISS---better retrieval quality with native explanations. \\
(b) & Overall Prox Imp & Average rating and rating variance drive global similarity---RFX-Fuse capability only. \\
\midrule
(c) & Toy Story Top-K & Most similar movies by tree-based proximity (Monsters Inc., Finding Nemo, etc.). \\
(d) & Toy Story Local Prox & Animation and Children genres drive this movie's neighborhood. \\
\midrule
(e) & The Matrix Top-K & Similar sci-fi/action movies ranked by proximity score. \\
(f) & The Matrix Local Prox & Sci-Fi and Action genres dominate---different drivers than Toy Story. \\
\bottomrule
\end{tabular}
\caption*{\textbf{Figure~\ref{fig:uc1-unsup} Panel Guide}}
\end{table}

\textbf{Key Insights (Part A: Unsupervised Retrieval):}
\begin{itemize}
    \item \textbf{Better hit rate and ranking}: HR@10 +25.4\% (0.454 vs 0.362), NDCG@10 +64.6\% (0.093 vs 0.057)---RFX-Fuse finds more relevant items and ranks them higher. Key advantage: native proximity importance explanations that FAISS cannot provide
    \item \textbf{Overall proximity importance} reveals global similarity drivers (importance scores in parentheses): avg\_rating (198.5), std\_rating (173.2), log\_users (103.8), log\_ratings (103.6), Drama (99.9)
    \item \textbf{Local proximity importance} explains \textit{why} each query's neighbors are similar---Toy Story: log\_ratings/log\_users/Animation (popular animated films); Matrix: log\_ratings/log\_users/Sci-Fi/Action (blockbuster sci-fi action)
    \item \textbf{No post-hoc computation}: All explanations computed during training (1,254s for 1,000 trees on 59K items); OOB Error=18.9\%
\end{itemize}

%--- FIGURE 2A: RFX-Fuse Supervised - Prediction + Similarity ---
\begin{figure}[H]
\centering
\includegraphics[width=\textwidth]{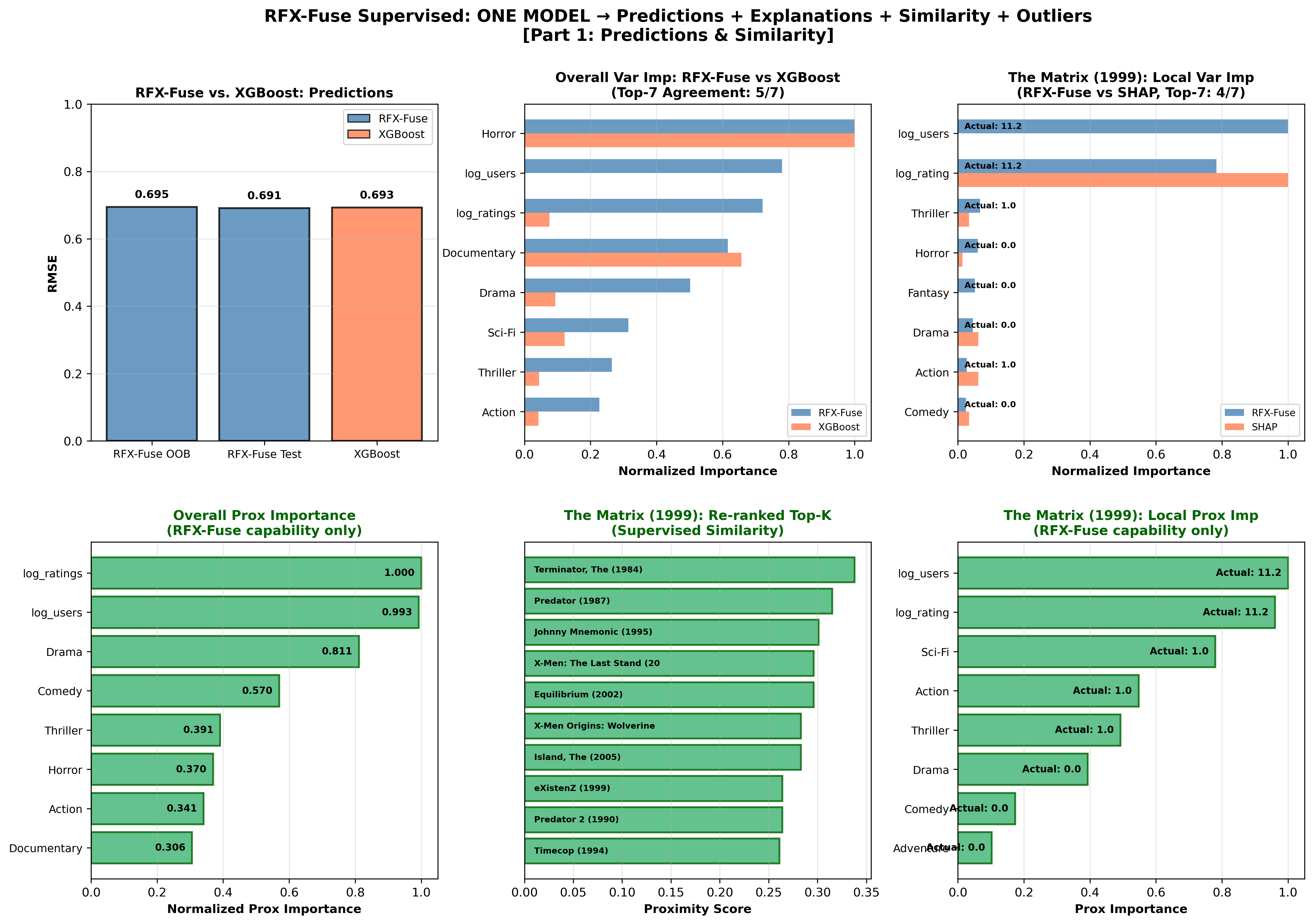}
\caption{RFX-Fuse Supervised (Part 1): ONE model provides predictions, explanations, similarity, and outlier detection---this figure shows predictions and similarity.}
\label{fig:uc1-sup-pred}
\end{figure}

\begin{table}[H]
\centering
\footnotesize
\begin{tabular}{@{}cl p{10.5cm}@{}}
\toprule
\textbf{Panel} & \textbf{Title} & \textbf{Description} \\
\midrule
(a--c) & Prediction & RFX-Fuse vs XGBoost RMSE, overall variable importance comparison, local variable importance vs SHAP for The Matrix. \\
(d--f) & Proximity & Overall proximity importance (RFX-Fuse capability only), re-ranked Top-K for The Matrix, local proximity importance (RFX-Fuse capability only). \\
\bottomrule
\end{tabular}
\caption*{\textbf{Figure~\ref{fig:uc1-sup-pred} Panel Guide}}
\end{table}

\textbf{Key Insights (Part B: Supervised Ranking):}
\begin{itemize}
    \item \textbf{Prediction accuracy}: RFX-Fuse Test (RMSE=0.691) matches XGBoost (RMSE=0.693); RFX-Fuse OOB (RMSE=0.695) matches both---OOB provides unbiased validation without held-out data (equivalent to $n$-tree cross-validation). RF with 1,000 trees cannot overfit; XGBoost required early stopping at 28 trees. Note: 1,000 trees used to demonstrate no-overfitting property; $\leq$200 trees typically sufficient for similar accuracy
    \item \textbf{Overall variable importance}: Horror (1.45), log\_users (1.13), log\_ratings (1.04) dominate; Top-10 agreement=20\% (different importance methods)
    \item \textbf{Local variable importance}: Top-10 agreement between RFX-Fuse and SHAP---both queries show 7/10 overlap; native explanations match post-hoc SHAP without additional computation
    \item \textbf{Local var imp vs local prox imp} (Matrix): Both share top-2 (log\_ratings, log\_users). Var imp ranks Drama \#3 (actual=0); prox imp ranks Sci-Fi \#3, Action \#4 (both=1). Insight: var imp captures what's \textit{absent} (not a drama); prox imp captures what's \textit{present} (clusters with sci-fi/action). Same model, complementary explanations
    \item \textbf{Re-ranking boost}: Supervised similarity achieves NDCG@10=0.66, HR@10=0.92---massive improvement over unsupervised (0.093, 0.454)
    \item \textbf{Single model}: Predictions + similarity + 4 importance types from ONE trained regressor (120.1s for 1,000 trees)
\end{itemize}

%--- FIGURE 2B: RFX-Fuse Supervised - Outlier Detection ---
\begin{figure}[H]
\centering
\includegraphics[width=\textwidth]{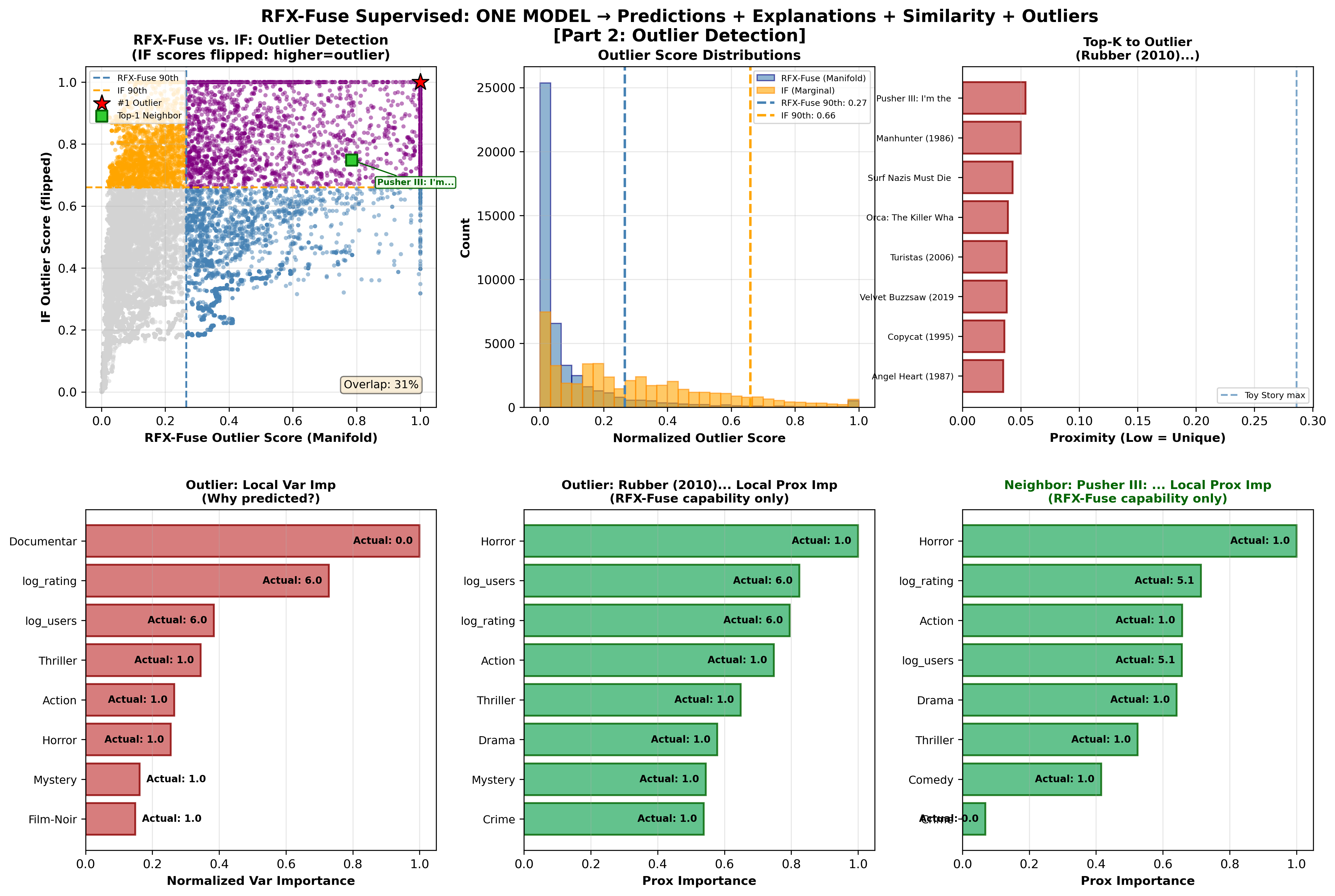}
\caption{RFX-Fuse Supervised (Part 2): ONE model provides predictions, explanations, similarity, and outlier detection---this figure shows outlier detection.}
\label{fig:uc1-sup-outlier}
\end{figure}

\begin{table}[H]
\centering
\footnotesize
\begin{tabular}{@{}cl p{10.5cm}@{}}
\toprule
\textbf{Panel} & \textbf{Title} & \textbf{Description} \\
\midrule
(a--c) & Outliers & RFX-Fuse vs Isolation Forest scatter plot, outlier score distributions, Top-K neighbors to top outlier (low proximity = unique). \textit{Note: IF scores flipped for comparison (RFX: higher=outlier; IF: lower=outlier).} \\
(d--f) & Outlier Analysis & Top outlier's local variable importance, outlier's local proximity importance (RFX-Fuse capability only), nearest neighbor's local proximity importance (RFX-Fuse capability only). \\
\bottomrule
\end{tabular}
\caption*{\textbf{Figure~\ref{fig:uc1-sup-outlier} Panel Guide}}
\end{table}

\textbf{Key Insights (Part B: Outlier Detection):}
\begin{itemize}
    \item \textbf{47.4\% overlap} (2,239/4,724) between RFX-Fuse and Isolation Forest outliers---methods detect \textit{different} anomaly types
    \item \textbf{Manifold vs marginal}: RFX-Fuse finds samples with low proximity to all clusters (unusual combinations); IF finds fast-split outliers (extreme individual values requiring few splits to isolate)
    \item \textbf{Outlier explanations}: RFX-Fuse provides local var imp (``why predicted?'') + local prox imp (``why isolated?'')---IF provides only scores
    \item \textbf{Top-K validation}: Outliers have uniformly low proximity scores to all neighbors, confirming isolation in manifold space
    \item \textbf{No additional model}: Outlier detection is built into the same regressor used for prediction and similarity
\end{itemize}

%--- FIGURE 3: Re-ranking Pipeline ---
\begin{figure}[H]
\centering
\includegraphics[width=\textwidth]{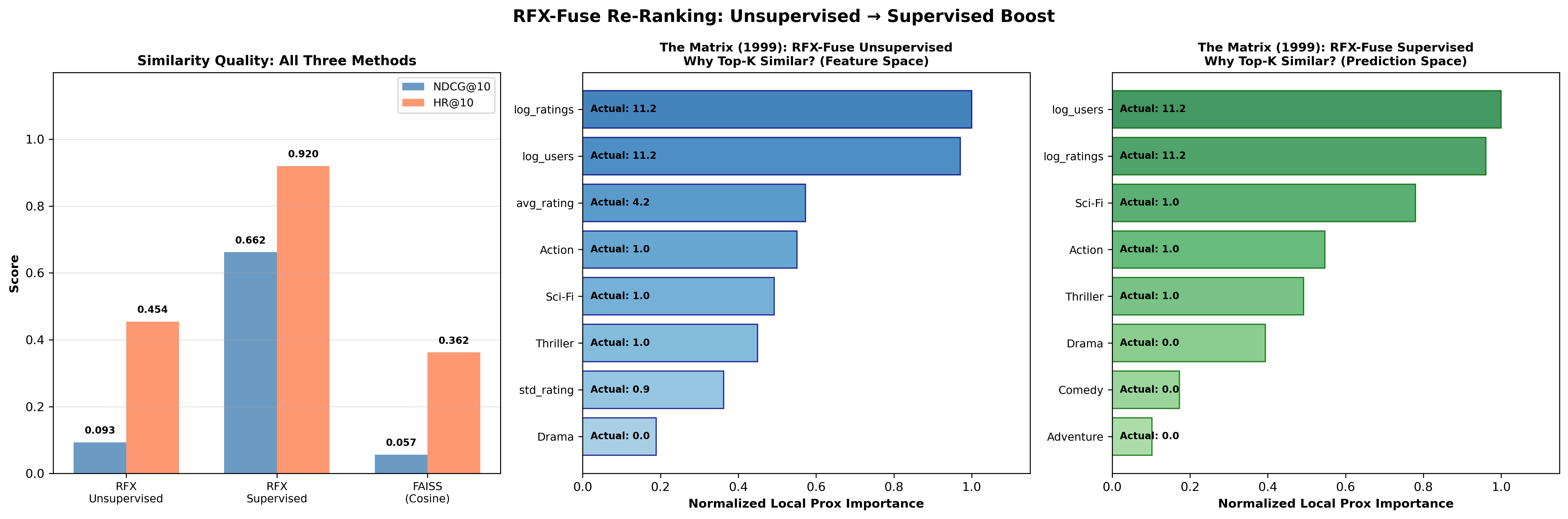}
\caption{RFX-Fuse re-ranking: Unsupervised $\rightarrow$ Supervised boost comparison.}
\label{fig:uc1-rerank}
\end{figure}

\begin{table}[H]
\centering
\footnotesize
\begin{tabular}{@{}cl p{10.5cm}@{}}
\toprule
\textbf{Panel} & \textbf{Title} & \textbf{Description} \\
\midrule
(a) & Three-Way Comparison & NDCG/HR metrics for RFX-Fuse Unsupervised, RFX-Fuse Supervised, and FAISS---supervised re-ranking improves upon unsupervised retrieval. \\
(b) & Unsup Local Prox Imp & The Matrix's similarity drivers in feature space with actual feature values shown (e.g., ``Sci-Fi (Actual: 1.0)'')---RFX-Fuse unique capability. \\
(c) & Sup Local Prox Imp & The Matrix's similarity drivers in prediction space with actual values---different features dominate based on learning objective. \\
\bottomrule
\end{tabular}
\caption*{\textbf{Figure~\ref{fig:uc1-rerank} Panel Guide}}
\end{table}

\textbf{Key Insights (Re-Ranking Pipeline):}
\begin{itemize}
    \item \textbf{Unsupervised $\rightarrow$ Supervised boost}: NDCG improves from 0.093 to 0.66 (+7$\times$); HR improves from 0.454 to 0.92 (+2$\times$). Implication: when content-based retrieval for similar features used in a re-ranker isn't required, a single supervised model provides a complete recommender system. Supervised top-K also serves as built-in re-ranker validation
    \item \textbf{Local prox imp comparison} (Matrix): Top-4 identical (log\_ratings, log\_users, Sci-Fi, Action). After top-4: Unsupervised includes avg\_rating (4.2), std\_rating (0.9)---raw rating statistics. Supervised includes Thriller (1), Drama (0), Comedy (0)---genre indicators only. Insight: Unsupervised finds ``similar content''; Supervised finds ``content that predicts similarly''---rating stats matter for raw similarity but genres drive prediction-space clustering
    \item \textbf{Complete pipeline}: RFX-Fuse offers a 2-model approach as an alternative to 5 tools (FAISS + XGBoost + SHAP + Isolation Forest + custom code)
\end{itemize}

\textbf{Use Case 1 Summary:} RFX-Fuse provides a complete recommender pipeline with native explainability. Two models (Unsupervised + Supervised) deliver retrieval, re-ranking, prediction, outlier detection, and four types of importance---all built into the model framework without external tools like SHAP or Isolation Forest. Alternatively, a single supervised model provides top-K similarity in prediction space, enabling a complete recommender system in as little as 1 model when content-based retrieval isn't required.

%==============================================================================
\subsection{Use Case 2: Finance Explainability}
%==============================================================================

\begin{table}[H]
\centering
\footnotesize
\begin{tabular}{@{}l p{12cm}@{}}
\toprule
\textbf{Aspect} & \textbf{Details} \\
\midrule
Problem & Loan denial prediction with regulatory-compliant explanations (ECOA, GDPR, Fair Lending). \\
Dataset & Kaggle Credit Score Classification (100K samples, 15 features, binary classification; 46K balanced train, 54K test). \\
\midrule
RFX-Fuse Pipeline & Single RFX-Fuse Classifier providing 4 importance types natively. \\
Industry Pipeline & XGBoost (prediction) + SHAP (local explanations) + Isolation Forest (outliers)---only 2 importance types, no similarity explanations. \\
\midrule
Evaluation & AUC-ROC, Precision/Recall, importance type coverage, regulatory compliance capability. \\
\bottomrule
\end{tabular}
\caption*{\textbf{Use Case 2 Summary}}
\end{table}

\begin{table}[H]
\centering
\footnotesize
\begin{tabular}{@{}l l p{9cm}@{}}
\toprule
\textbf{Stage} & \textbf{Comparison} & \textbf{Evaluation Method} \\
\midrule
\multicolumn{3}{l}{\textit{Prediction Performance}} \\
\midrule
Classification & RFX-Fuse vs XGBoost & AUC-ROC, Precision, Recall, F1 on held-out test set. \\
Calibration & RFX-Fuse vs XGBoost & ROC and PR curve comparison. \\
\midrule
\multicolumn{3}{l}{\textit{Explainability (4 Types)}} \\
\midrule
Overall Var Imp & RFX-Fuse vs XGBoost & Global feature rankings for default prediction. \\
Local Var Imp & RFX-Fuse vs SHAP & Per-sample feature contributions (``why denied?''). \\
Overall Prox Imp & RFX-Fuse only & Global similarity structure (``what clusters borrowers?''). \\
Local Prox Imp & RFX-Fuse only & Per-sample similarity explanation (``why similar?''). \\
\midrule
\multicolumn{3}{l}{\textit{Outlier Detection}} \\
\midrule
Detection & RFX-Fuse vs IF & Score correlation, outlier overlap, interpretability. \\
Analysis & RFX-Fuse only & Top-K similar to outlier + proximity importance explanation. \\
\bottomrule
\end{tabular}
\caption*{\textbf{Evaluation Methodology}}
\end{table}

%--- FIGURE 2A: Finance Part A ---
\begin{figure}[H]
\centering
\includegraphics[width=\textwidth]{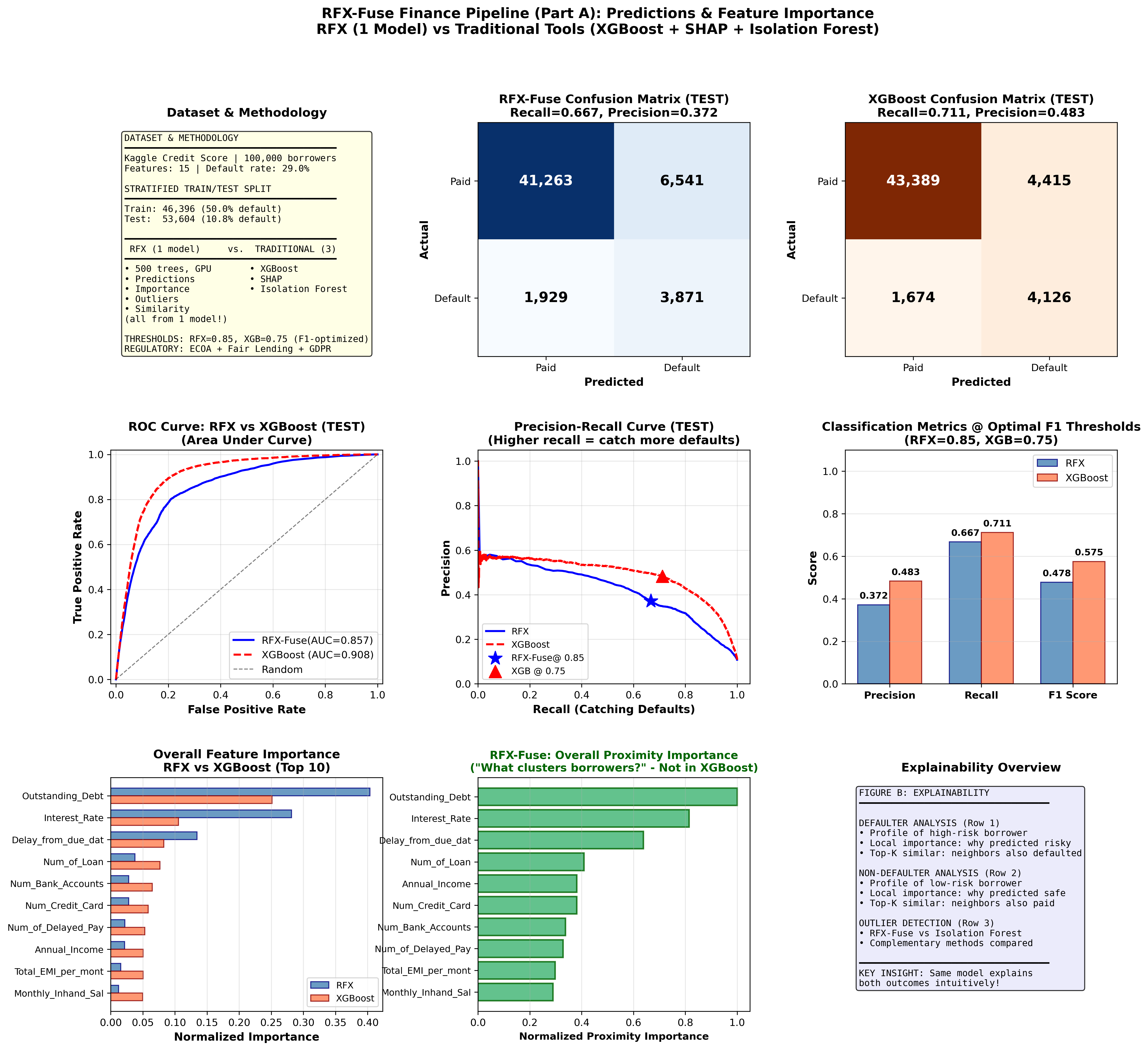}
\caption{\textbf{Use Case 2: Finance Explainability (Part A).} Predictions and feature importance.}
\label{fig:usecase2a}
\end{figure}

\begin{table}[H]
\centering
\footnotesize
\begin{tabular}{@{}cl p{9cm}@{}}
\toprule
\textbf{Panel} & \textbf{Title} & \textbf{Description} \\
\midrule
(a) & Dataset & Kaggle Credit Score: 100K samples, 15 features; 46K balanced train (50\% default), 54K test (10.8\% default). \\
(b,c) & Confusion Matrices & RFX-Fuse vs XGBoost test set performance---comparable recall/precision. \\
(d,e) & ROC \& PR Curves & AUC comparison showing RFX-Fuse matches XGBoost predictive performance. \\
(f) & Metrics Bar Chart & Side-by-side metrics at optimized threshold for recall. \\
(g) & Overall Var Imp & RFX-Fuse vs XGBoost global feature rankings---both identify key default predictors. \\
(h) & Overall Prox Imp & What clusters borrowers globally (RFX-Fuse capability only). \\
(i) & Summary & Overview of explainability capabilities in Figure B. \\
\bottomrule
\end{tabular}
\caption*{\textbf{Figure~\ref{fig:usecase2a} Panel Guide}}
\end{table}

\textbf{Key Insights (Part A: Predictions \& Global Importance):}
\begin{itemize}
    \item \textbf{Recall-optimized thresholds}: RFX-Fuse (thresh=0.85): Recall=0.667, F1=0.478; XGBoost (thresh=0.75): Recall=0.711, F1=0.575---both tuned for catching defaults
    \item \textbf{Prediction parity}: RFX-Fuse 81.1\% vs XGBoost 84.0\% accuracy (2.9\% gap)---minor trade-off for 4$\times$ more explanation types; OOB Error=21.2\% provides built-in validation
    \item \textbf{Overall var imp agreement}: Top-5 identical across methods---Outstanding\_Debt (RFX: 0.41, XGB: 0.25), Interest\_Rate (0.27 vs 0.11), Delay\_from\_due\_date (0.13 vs 0.08)
    \item \textbf{Overall prox imp vs var imp}: Same top features, different rankings (prox imp scores in parentheses)---Outstanding\_Debt (242.4), Interest\_Rate (197.5), Delay\_from\_due\_date (154.7) cluster borrowers; Num\_of\_Loan ranks 4th in prox imp (99.1)
    \item \textbf{Unique insight}: Num\_Credit\_Card clusters borrowers strongly (prox imp=92.5) but predicts default weakly---indicates credit behavior segments, useful for fair lending analysis
    \item \textbf{Training time}: 69.0s for 500 trees on 46K balanced samples---all 4 importance types computed natively
\end{itemize}

%--- FIGURE 2B: Finance Part B ---
\begin{figure}[H]
\centering
\includegraphics[width=\textwidth]{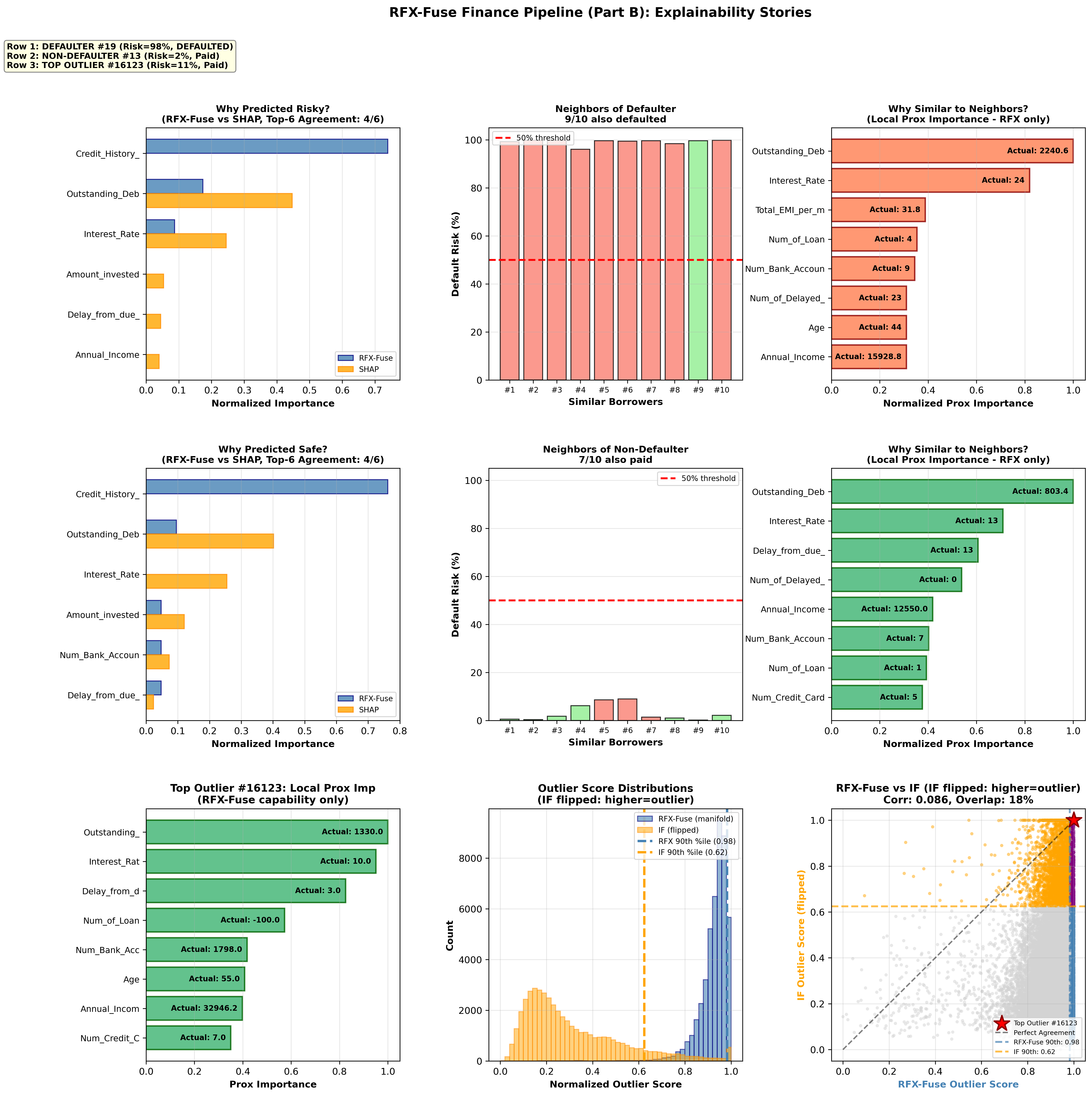}
\caption{\textbf{Use Case 2: Finance Explainability (Part B).} Individual explainability stories for three borrower types.}
\label{fig:usecase2b}
\end{figure}

\begin{table}[H]
\centering
\footnotesize
\begin{tabular}{@{}cl p{9cm}@{}}
\toprule
\textbf{Panel} & \textbf{Title} & \textbf{Description} \\
\midrule
\multicolumn{3}{l}{\textit{Row 1: Defaulter Analysis}} \\
\midrule
(j) & RFX vs SHAP & Why this borrower was predicted risky (Top-6 Agreement shows feature consistency). \\
(k) & Neighbors & Top-K similar borrowers---validates prediction via peer outcomes. \\
(l) & Local Prox Imp & Why similar to neighbors? Explains clustering drivers (RFX-Fuse only). \\
\midrule
\multicolumn{3}{l}{\textit{Row 2: Non-Defaulter Analysis}} \\
\midrule
(m) & RFX vs SHAP & Why this borrower was predicted safe (Top-6 Agreement shows feature consistency). \\
(n) & Neighbors & Top-K similar borrowers---validates prediction via peer outcomes. \\
(o) & Local Prox Imp & Why similar to neighbors? Different drivers than defaulter (RFX-Fuse only). \\
\midrule
\multicolumn{3}{l}{\textit{Row 3: Outlier Analysis}} \\
\midrule
(p) & Local Prox Imp & What makes this outlier unusual? Feature values driving isolation (RFX-Fuse only). \\
(q) & Distributions & RFX-Fuse (manifold) vs IF (marginal) outlier score distributions. \textit{Note: IF scores flipped for comparison (RFX: higher=outlier; IF: lower=outlier).} \\
(r) & RFX vs IF Scatter & Correlation between detection methods; different anomaly types identified. \textit{IF flipped for comparison.} \\
\bottomrule
\end{tabular}
\caption*{\textbf{Figure~\ref{fig:usecase2b} Panel Guide}}
\end{table}

\textbf{Key Insights (Row 1: Defaulter \#19):}
\begin{itemize}
    \item \textbf{High-risk prediction}: P(Default)=98.4\%---model confident this borrower will default
    \item \textbf{Neighbor validation}: 9/10 similar borrowers in feature space also defaulted---strong peer evidence confirms high-risk prediction
    \item \textbf{Local prox imp drivers}: Outstanding\_Debt=\$2,241 (0.0068), Interest\_Rate=24\% (0.0056), Total\_EMI=\$32 (0.0026)---these features drive similarity clustering
    \item \textbf{Adverse action notice}: ``Denied due to outstanding debt (\$2,241) and elevated interest rate (24\%). Credit\_History\_Age=0 is primary local var imp factor.''
\end{itemize}

\textbf{Key Insights (Row 2: Non-Defaulter \#13):}
\begin{itemize}
    \item \textbf{Low-risk prediction}: P(Default)=2.4\%---model confident this borrower will pay
    \item \textbf{Neighbor validation}: 7/10 similar borrowers paid successfully---strong peer evidence supports safe prediction
    \item \textbf{Protective factors}: Outstanding\_Debt=\$803 (0.0069), Interest\_Rate=13\% (0.0049), Delay\_from\_due\_date=13 days (0.0042), Num\_of\_Delayed\_Payment=0 (0.0037)
    \item \textbf{Contrastive insight}: Defaulter has Outstanding\_Debt=\$2,241 vs Non-Defaulter=\$803; Interest\_Rate=24\% vs 13\%---clear decision boundary
\end{itemize}

\textbf{Key Insights (Row 3: Top Outlier \#16123):}
\begin{itemize}
    \item \textbf{Unusual profile}: P(Default)=5\% yet flagged as outlier---unusual combination of features makes this borrower unlike any cluster
    \item \textbf{RFX vs Isolation Forest}: Different detection approaches---RFX finds manifold outliers (unusual combinations); IF flags extreme individual values
    \item \textbf{Actionable triage}: Local prox imp explains \textit{why} flagged---enables efficient manual review for fraud/risk teams
\end{itemize}

\textbf{Use Case 2 Summary:}
\begin{itemize}
    \item \textbf{Regulatory compliance}: RFX-Fuse answers both ``why denied?'' (local var imp) AND ``who else was denied similarly?'' (neighbors + local prox imp)---required for ECOA, GDPR Article 22, Fair Lending
    \item \textbf{Single model}: 1 RFX-Fuse classifier provides a comparable alternative to XGBoost + SHAP + Isolation Forest with native 4-type explainability
    \item \textbf{No post-hoc computation}: All explanations computed during training---production-ready without SHAP overhead
\end{itemize}

%==============================================================================
\subsection{Use Case 3: Time-Series Regression}
%==============================================================================

\begin{table}[H]
\centering
\footnotesize
\begin{tabular}{@{}l p{12cm}@{}}
\toprule
\textbf{Aspect} & \textbf{Details} \\
\midrule
Problem & Hourly bike rental prediction with anomaly detection and multi-type explanations. \\
Dataset & UCI Bike Sharing~\cite{bikesharing2013} (8,645 hourly records from 2011, 4 core features: hr, season, mnth, temp). Train: 5,725 (Jan--Aug), Test: 2,920 (Sep--Dec). \\
\midrule
RFX-Fuse Pipeline & Single RFX-Fuse Regressor providing prediction + 4 importance types + outlier detection. \\
Industry Pipeline & XGBoost (prediction) + SHAP (explanations) + Isolation Forest (outliers) + FAISS (similarity)---4 tools. \\
\midrule
Evaluation & RMSE/MSE on test set, importance ranking comparison, outlier score correlation. \\
\bottomrule
\end{tabular}
\caption*{\textbf{Use Case 3 Summary}}
\end{table}

\begin{table}[H]
\centering
\footnotesize
\begin{tabular}{@{}l l p{9cm}@{}}
\toprule
\textbf{Stage} & \textbf{Comparison} & \textbf{Evaluation Method} \\
\midrule
\multicolumn{3}{l}{\textit{Prediction Performance}} \\
\midrule
Regression & RFX-Fuse vs XGBoost & RMSE/MSE on held-out test set (last month of data). \\
Time Series Fit & RFX-Fuse vs XGBoost & Visual comparison of predicted vs actual hourly rentals. \\
\midrule
\multicolumn{3}{l}{\textit{Explainability}} \\
\midrule
Overall Var Imp & RFX-Fuse vs XGBoost & Global feature rankings (hour, temperature, month, season). \\
Local Var Imp & RFX-Fuse vs SHAP & Per-sample feature contributions for predictions. \\
Overall Prox Imp & RFX-Fuse only & Global similarity structure in rental patterns. \\
Local Prox Imp & RFX-Fuse only & Per-sample similarity explanation. \\
\midrule
\multicolumn{3}{l}{\textit{Outlier Detection}} \\
\midrule
Detection & RFX-Fuse vs IF & Score correlation, distribution comparison, outlier overlap. \\
Analysis & RFX-Fuse only & Top-K similar + local var/prox importance for outliers. \\
\bottomrule
\end{tabular}
\caption*{\textbf{Evaluation Methodology}}
\end{table}

\begin{figure}[H]
\centering
\includegraphics[width=\textwidth]{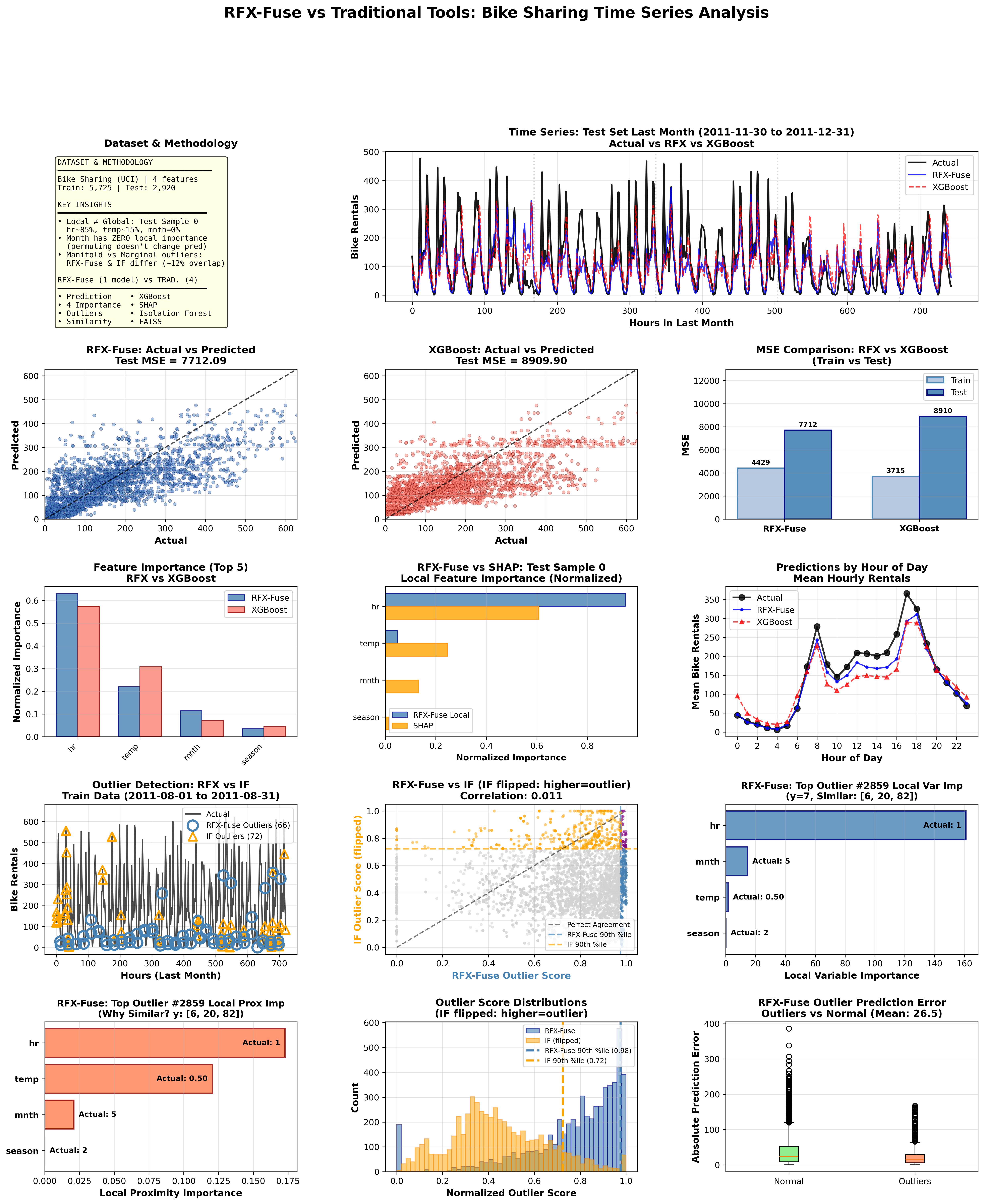}
\caption{\textbf{Use Case 3: Bike Sharing Regression.} RFX-Fuse provides prediction, variable importance, proximity importance, and outlier detection in a single model. Traditional approach requires XGBoost (prediction) + SHAP (explanations) + Isolation Forest (outliers) + FAISS (similarity)---4 separate tools.}
\label{fig:usecase3}
\end{figure}

\begin{table}[H]
\centering
\footnotesize
\begin{tabular}{@{}cl p{10cm}@{}}
\toprule
\textbf{Panel} & \textbf{Title} & \textbf{Description} \\
\midrule
(a) & Time Series & Test set predictions (last month): actual vs RFX-Fuse vs XGBoost hourly rentals. \\
(b,c) & Actual vs Predicted & Scatter plots showing RFX-Fuse and XGBoost test set fit quality. \\
\midrule
(d) & MSE Comparison & Train vs test MSE---RFX-Fuse (7,671) outperforms XGBoost (10,203) on test set. \\
(e) & Feature Importance & RFX-Fuse vs XGBoost global rankings---hour, temperature, month, season. \\
(f) & RFX-Fuse vs SHAP & Local variable importance comparison for a test sample. \\
\midrule
(g) & Predictions by Hour & Mean hourly rentals showing daily usage patterns captured by both models. \\
(h) & Outlier Detection TS & Time series with outliers highlighted (unusual rental patterns). \\
(i) & RFX-Fuse vs IF & Scatter comparing RFX-Fuse and Isolation Forest outlier scores. \textit{IF flipped for comparison.} \\
\midrule
(j) & Top Outlier Var Imp & Why the top outlier was predicted this way. \\
(k) & Top Outlier Prox Imp & Why the outlier's neighbors are similar (RFX-Fuse capability only). \\
(l) & Outlier Distributions & RFX-Fuse (manifold) vs IF (marginal) score histograms. \textit{Note: IF scores flipped (RFX: higher=outlier; IF: lower=outlier).} \\
\midrule
(m--o) & Outlier Analysis & Prediction error analysis for outliers vs normal samples. \\
\bottomrule
\end{tabular}
\caption*{\textbf{Figure~\ref{fig:usecase3} Panel Guide}}
\end{table}

\textbf{Key Results:}
\begin{itemize}
    \item \textbf{Prediction:} RFX-Fuse (RMSE=87.8) outperforms XGBoost (RMSE=94.4) on this time-series regression task, with equivalent overall feature importance rankings (hour $>$ temp $>$ month $>$ season). OOB RMSE=66.6 provides built-in validation without a held-out set.

    \item \textbf{Local vs Global importance:} Global importance shows which features matter across all samples. Local importance reveals that for a sample, \texttt{hour} dominates ($\sim$99.7\%), \texttt{temp} contributes $\sim$0.3\%, while \texttt{season} and \texttt{month} have \textit{zero} local importance---permuting them doesn't change this sample's prediction. This sample-specific insight is unavailable from global importance alone.

    \item \textbf{Explainability equivalence:} XGBoost + SHAP provides overall variable importance (global) + local variable importance (per-sample). RFX-Fuse provides \textit{both} natively, \textit{plus} overall proximity importance (global similarity structure) + local proximity importance (per-sample similarity explanation)---4 importance types vs 2.

    \item \textbf{Outlier detection---manifold vs marginal:} RFX-Fuse and Isolation Forest identify different outliers ($\sim$10\% overlap). Why? IF detects \textit{marginal} outliers via short tree paths---samples with extreme feature values split quickly in random trees. RFX-Fuse detects \textit{manifold} outliers---samples that don't fit the learned feature interactions. Top outliers are unusual in feature space but still predictable, revealing latent structure IF misses. Compare to Use Case 1's 47\% overlap: low overlap ($\sim$10\%) indicates complex multivariate patterns where RFX finds manifold outliers IF misses; high overlap ($\sim$47\%) indicates simpler feature interactions where methods converge. The overlap percentage itself is a diagnostic for dataset complexity.

    \item \textbf{Why similar?} For any outlier, RFX-Fuse provides: (1) local variable importance explaining \textit{why predicted this way}, and (2) local proximity importance explaining \textit{why similar to neighbors}---two complementary explanations unavailable in traditional pipelines.
\end{itemize}

%==============================================================================
\subsection{Use Case 4: Imputation Quality Validation}
%==============================================================================

\noindent\textbf{The Problem:} How do you validate imputation quality \textit{without ground truth}? If we had ground truth, we wouldn't need imputation. Traditional approaches (MSE, cross-validation) require known true values---a circular dependency.

\vspace{0.5em}
\noindent\textbf{RFX-Fuse Solution:} Train RFX-Fuse Unsupervised on original (complete) data. The model learns to distinguish ``real'' data from ``synthetic'' permuted data. Poorly imputed data distorts the original feature dependencies and looks more ``synthetic''---higher P(synthetic) = worse imputation. For time series, prior work~\cite{farjallah2025} uses statistical divergence metrics (Wasserstein distance, Jensen-Shannon divergence) for temporal distribution matching. For general tabular data, RFX-Fuse provides a learning-based approach that captures \textit{multivariate dependency structure} via learned feature interactions.

\vspace{0.5em}
\noindent\textbf{Dataset:} UCI Bike Sharing (3,000 hourly records, 12 features). We introduce 15\% missing values and compare 5 imputation methods.

\begin{figure}[H]
\centering
\includegraphics[width=\textwidth]{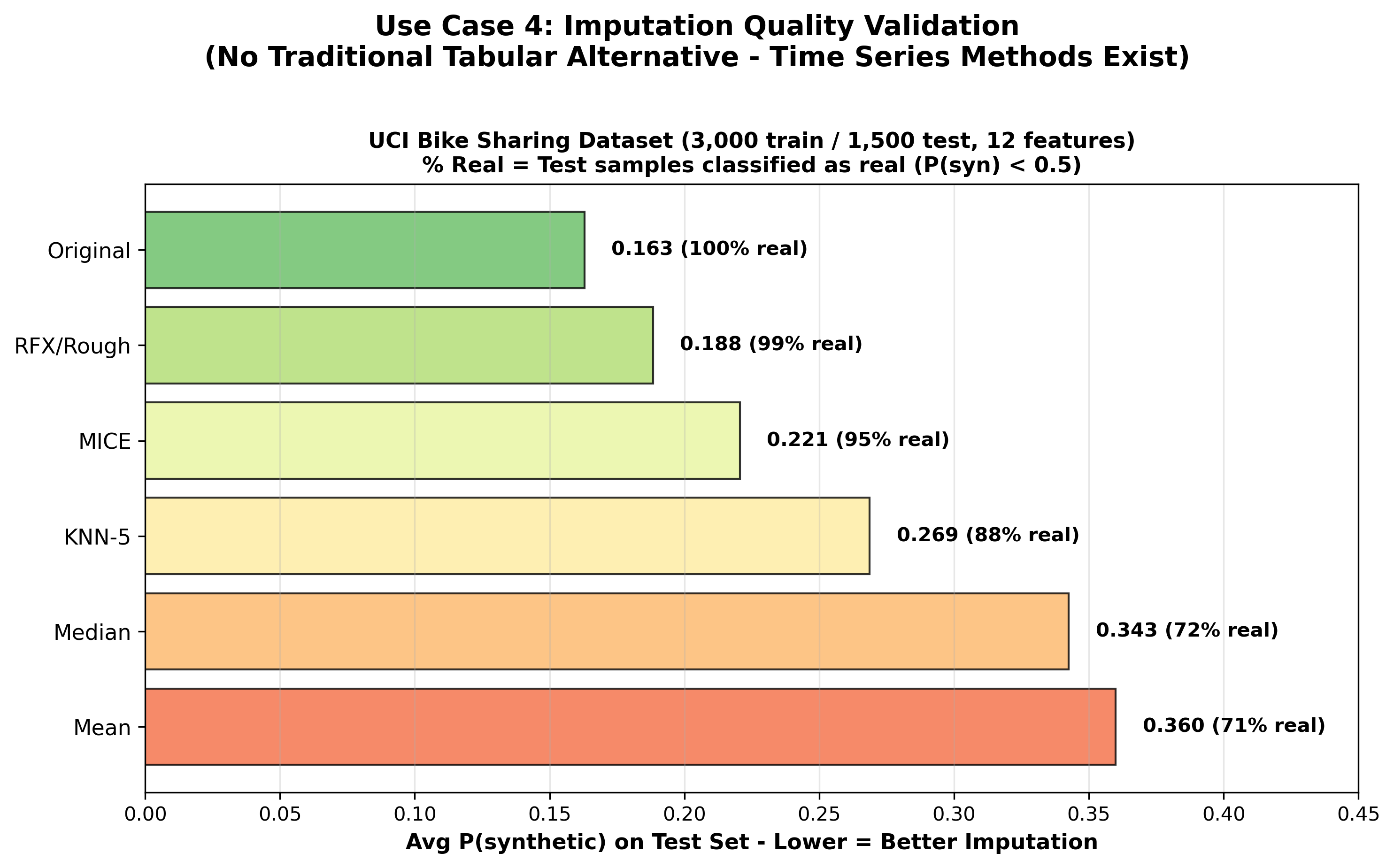}
\caption{\textbf{Use Case 4: Imputation Quality Validation.} RFX-Fuse Unsupervised ranks imputation methods by P(synthetic)---lower means data looks more ``real.'' For general tabular data, RFX-Fuse captures multivariate dependency structure via learned feature interactions.}
\label{fig:usecase4}
\end{figure}

\textbf{Key Results:}
\begin{itemize}
    \item \textbf{Imputation ranking:} Original=100\% $>$ RFX/Rough=98\% $>$ MICE=95\% $>$ KNN=88\% $>$ Median=72\% $>$ Mean=70\%

    \item \textbf{Structure-preserving wins:} Methods that preserve feature correlations (RFX/Rough, MICE) score higher P(real) than simple methods (Mean, Median) that distort the data structure.

    \item \textbf{Novel capability:} For time series, distribution-matching metrics exist~\cite{farjallah2025}. For general tabular data, RFX-Fuse Unsupervised provides a learning-based approach---ranking imputation methods by how well they preserve multivariate feature interactions.
\end{itemize}

%==============================================================================
\subsection{Use Case 5: Anomaly Detection (Breiman-Cutler Method)}
%==============================================================================

This use case implements Breiman and Cutler's original vision for anomaly detection: train on clean data, then identify samples that don't fit the learned data manifold as outliers. This is fundamentally different from Isolation Forest's marginal approach, which scores outliers based on short tree paths---samples with extreme feature values are isolated quickly in random splits, requiring fewer splits to separate them from the data.

\begin{table}[H]
\centering
\footnotesize
\begin{tabular}{@{}l p{12cm}@{}}
\toprule
\textbf{Aspect} & \textbf{Details} \\
\midrule
Problem & Detect anomalies in financial data without labeled examples. \\
Dataset & Kaggle Credit Score (15K clean training, 5K test with 50\% injected anomalies: point, contextual, collective). \\
\midrule
RFX-Fuse Pipeline & Train Unsupervised on \textit{clean data only} $\rightarrow$ Anomalies have high P(synthetic). \\
Industry Pipeline & Isolation Forest (marginal detection, no training/test split, no explanations). \\
\midrule
Evaluation & AUC-ROC, Precision@K, P(synthetic) separation between clean and anomaly samples. \\
\bottomrule
\end{tabular}
\caption*{\textbf{Use Case 5 Summary}}
\end{table}

\begin{table}[H]
\centering
\footnotesize
\begin{tabular}{@{}l l p{9cm}@{}}
\toprule
\textbf{Stage} & \textbf{Comparison} & \textbf{Evaluation Method} \\
\midrule
\multicolumn{3}{l}{\textit{Global Anomaly Detection}} \\
\midrule
Detection & RFX-Fuse vs IF & AUC-ROC on held-out test set with injected anomalies. \\
Score Separation & RFX-Fuse vs IF & P(synthetic) distribution separation between clean and anomaly samples. \\
Feature Importance & RFX-Fuse unique & Overall feature importance for anomaly detection (complements IF's fast detection). \\
\midrule
\multicolumn{3}{l}{\textit{Individual Outlier Analysis (RFX-Fuse unique capabilities)}} \\
\midrule
Top-K Similar & RFX-Fuse unique & Find most similar samples to the top outlier via proximity matrix. \\
Local Var Imp & RFX-Fuse unique & Explain \textit{why} the outlier was classified this way (per-feature contribution). \\
Local Prox Imp & RFX-Fuse unique & Explain \textit{why} the outlier is unique (which features distinguish it from neighbors). \\
\bottomrule
\end{tabular}
\caption*{\textbf{Evaluation Methodology}}
\end{table}

\begin{figure}[H]
\centering
\includegraphics[width=\textwidth]{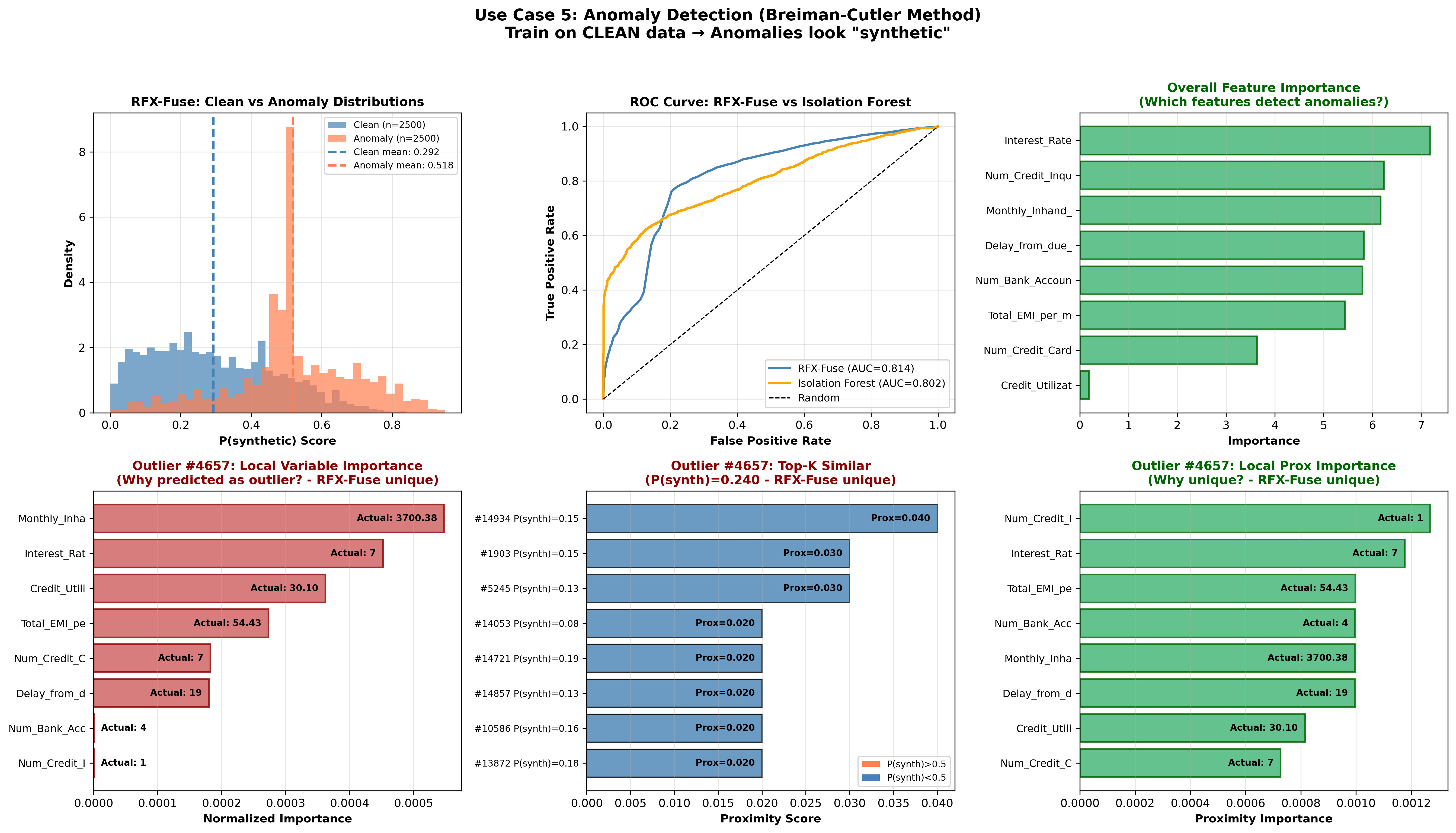}
\caption{\textbf{Use Case 5: Anomaly Detection (Breiman-Cutler Method).} RFX-Fuse Unsupervised trained on clean data assigns higher P(synthetic) to anomalies. This implements Breiman \& Cutler's original insight: anomalies don't fit the learned data manifold.}
\label{fig:usecase5}
\end{figure}

\begin{table}[H]
\centering
\footnotesize
\begin{tabular}{@{}cl p{10cm}@{}}
\toprule
\textbf{Panel} & \textbf{Title} & \textbf{Description} \\
\midrule
(a) & Score Distributions & Clean samples have low P(synthetic) ($\mu$=0.29); anomalies have high P(synthetic) ($\mu$=0.52)---clear separation validating Breiman-Cutler method. \\
(b) & ROC Curves & RFX-Fuse (AUC=0.81) outperforms Isolation Forest (AUC=0.80)---manifold-aware detection. \\
(c) & Overall Var Importance & Which features distinguish real from synthetic data globally (RFX-Fuse unique). \\
\midrule
(d) & Local Var Importance & Per-feature contributions explaining \textit{why} the top outlier was predicted as anomaly (RFX-Fuse unique). \\
(e) & Top Outlier Top-K & ``Why unique?''---top-K shows uniformly \textit{low} proximity scores to all neighbors because this is a true manifold outlier isolated from all other samples (RFX-Fuse unique). \\
(f) & Local Prox Importance & Which features make the outlier \textit{unique}---what it shares with its nearest neighbors (very little), explaining its isolation (RFX-Fuse unique). \\
\bottomrule
\end{tabular}
\caption*{\textbf{Figure~\ref{fig:usecase5} Panel Guide}}
\end{table}

\textbf{Key Results:}
\begin{itemize}
    \item \textbf{Breiman-Cutler insight}: Train on clean data $\rightarrow$ anomalies look ``synthetic'' (high P(synthetic))
    \item RFX-Fuse AUC=0.81 vs Isolation Forest AUC=0.80---manifold-aware detection comparable to specialized tools
    \item Clean samples: P(synthetic)=0.29; Anomalies: P(synthetic)=0.52 (separation=0.23)---clear distributional separation
    \item \textbf{Overall var importance}: Interest\_Rate (7.19), Num\_Credit\_Inquiries (6.24), Monthly\_Inhand\_Salary (6.17) distinguish real from synthetic data globally
    \item \textbf{``Why unique?'' (Top-K)}: Outlier \#4657 (P(syn)=0.24) has uniformly low proximity to all neighbors (avg=0.025)---isolated in manifold space. Top-8 neighbors all have proximity $\leq$0.04 and P(syn)$<$0.20 (normal samples)
    \item \textbf{Local var importance}: Outlier \#4657 driven by Monthly\_Inhand\_Salary=\$3,700 (0.00055), Interest\_Rate=7\% (0.00045), Credit\_Utilization=30\% (0.00036)---explains \textit{why predicted as anomaly}
    \item \textbf{Local prox importance}: Num\_Credit\_Inquiries (0.00127), Interest\_Rate (0.00118), Total\_EMI (0.00100)---explains \textit{what makes it unique} (features shared with neighbors, very little for true outliers)
\end{itemize}

%==============================================================================
\section{Conclusion}
%==============================================================================

This paper presented RFX-Fuse, completing Breiman and Cutler's original vision for Random Forests as a unified ML engine. The key novel contribution is Proximity Importance---native explainable similarity that explains \textit{why} samples are similar, not just \textit{that} they are similar. We demonstrate RFX-Fuse as a comparable alternative to modern multi-tool pipelines (XGBoost, FAISS, SHAP, Isolation Forest) across five industry use cases:

\begin{itemize}
    \item \textbf{Recommender Systems:} Two-stage retrieval $\rightarrow$ ranking pipeline with native re-ranking validation---supervised top-K similarity provides built-in candidate re-ranking without additional infrastructure
    \item \textbf{Finance Explainability:} Complete 4-type importance for regulatory compliance (ECOA, GDPR, Fair Lending)---``why denied?'' and ``who else was denied similarly?'' from one model
    \item \textbf{Time-Series Regression:} 1 model as comparable alternative to 4 tools (XGBoost + SHAP + Isolation Forest + FAISS)
    \item \textbf{Imputation Validation:} Rank imputation methods without ground truth via learned multivariate dependencies
    \item \textbf{Anomaly Detection:} Breiman-Cutler outlier detection with built-in explanations (why outlier?)
\end{itemize}

RFX-Fuse is a unified learning and similarity engine built on Breiman and Cutler's complete Random Forest vision. A single model provides prediction + prediction importance + similarity + similarity importance in a production-ready, GPU-accelerated library. By answering both ``why predicted?'' and ``why similar?'' from one framework, RFX-Fuse offers a streamlined alternative to multi-tool pipelines for teams seeking explainable ML with built-in similarity capabilities.

\textbf{Code:} \url{https://github.com/chriskuchar/RFX-Fuse}

%==============================================================================
\section*{Acknowledgements}
%==============================================================================

The author extends sincere thanks to Dr. Adele Cutler for generously sharing original Breiman-Cutler Random Forest source materials, which made this faithful restoration and extension possible. Her trust and support were invaluable in bringing Breiman and Cutler's original vision to modern infrastructure. The author also thanks his wife for her patience and support throughout this project.

%==============================================================================

\end{document}